\documentclass[conference]{IEEEtran}  % or use other templates like \documentclass[journal]{IEEEtran}

\usepackage[utf8]{inputenc}
\usepackage{graphicx}
\usepackage{amsmath}
\usepackage{amssymb}
\usepackage{cite}
\usepackage{url}
\usepackage{hyperref}
\usepackage{booktabs}
\usepackage{siunitx}

\title{Incorporating Error Level Noise Embedding for Improving LLM-Assisted Robustness in Persian Speech Recognition}

\author{
    \IEEEauthorblockN{Zahra Rahmani}
    \IEEEauthorblockA{Department of Computer Engineering \\
    Sharif University of Technology \\
    Email: zahra.rahmaniez@gmail.com}
    \and
    \IEEEauthorblockN{Hossein Sameti}
    \IEEEauthorblockA{Department of Computer Engineering \\
    Sharif University of Technology \\
    Email: sameti@sharif.edu}
}

\begin{document}

\maketitle

\begin{abstract}
Automatic Speech Recognition (ASR) systems suffer significant performance degradation in noisy environments, a challenge that is especially severe for low-resource languages such as Persian. Even state-of-the-art models such as Whisper struggle to maintain accuracy under varying signal-to-noise ratios (SNRs). This study presents a robust noise-sensitive ASR error correction framework that combines multiple hypotheses and noise-aware modeling. Using noisy Persian speech, we generate 5-best hypotheses from a modified Whisper-large decoder. Error Level Noise (ELN) is introduced as a representation that captures semantic- and token-level disagreement across hypotheses, quantifying the linguistic distortions caused by noise. ELN thus provides a direct measure of noise-induced uncertainty, enabling the LLM to reason about the reliability of each hypothesis during correction. Three models are evaluated: (1) a base LLaMA-2-7B model without fine-tuning, (2) a fine-tuned variant trained on text-only hypotheses, and (3) a noise-conditioned model integrating ELN embeddings at both sentence and word levels. Experimental results demonstrate that the ELN-conditioned model achieves substantial reductions in Word Error Rate (WER). Specifically, on the challenging Mixed Noise test set, the proposed Fine-tuned + ELN (Ours) model reduces the WER from a baseline of 31.10\% (Raw Whisper) to 24.84\%, significantly surpassing the Fine-tuned (No ELN) text-only baseline of 30.79\%, whereas the original LLaMA-2-7B model increased the WER to 64.58\%, demonstrating that it is unable to correct Persian errors on its own. This confirms the effectiveness of combining multiple hypotheses with noise-aware embeddings for robust Persian ASR in noisy real-world scenarios.
\end{abstract}


\section{Introduction}

Automatic Speech Recognition has become an integral part in the interaction of humans with computers in the modern world, and it finds applications in smart assistants, speech translation, and automatic captioning. The ever-increasing integration of ASR systems into everyday technology underscores the requirement for robustness and reliability across a wide variety of acoustic conditions and languages. However, while recent advances in end-to-end architectures and large language models have dramatically improved the performance of ASR in high-resource languages such as English and Mandarin, maintaining accuracy under noisy environments and for low-resource languages remains an open challenge.

Several lines of prior research have been pursued toward enhancing ASR robustness, including noise-aware training, data augmentation with synthetic noise, and post-processing using language models or sequence correction networks.
Noise-aware and multi-condition training techniques have long been effective in improving ASR robustness to acoustic distortions~\cite{ko2015audio, watanabe2018espnet}, while data augmentation methods such as reverberation and additive noise mixing remain standard practice for robustness enhancement~\cite{li2016robust}.
Despite their effectiveness, these approaches operate primarily at the acoustic modeling level and often require retraining or fine-tuning the ASR model itself. 
While these methods have achieved certain improvements, they still suffer from several limitations: most approaches are developed and evaluated for high-resource languages, leaving low-resource languages like Persian underrepresented; existing correction models usually work only on the textual outputs of ASR, without considering the noise information that leads to transcription errors; many systems try to modify or retrain the original ASR architecture, which is computationally expensive and limits adaptability; and few prior methods explicitly explore large generative models or multi-hypothesis reasoning, most of which are not noise-robust, for leveraging contextual and semantic cues for error correction.

This study addresses these gaps by introducing the first noise-robust generative error correction model for Persian ASR. Unlike previous approaches, ours focuses on post-hoc correction and is lightweight and easily deployable without requiring any retraining of the ASR model. Specifically, we generate multiple transcription hypotheses (5-best) from a Whisper-based Persian ASR system and design a noise-conditioned correction mechanism using a large language model, LLaMA-2-7B~\cite{touvron2023llama2}. We want to make the model aware of the acoustic degradation affecting each utterance. So, we develop Error Level Noise (ELN) embeddings that encode sentence- and token-level disagreement across hypotheses as a proxy for noise impact. These are used in order to condition the LLM during fine-tuning, allowing it to integrate both linguistic and noise-aware cues during correction. Our experiments on Persian speech, which originated from Common Voice~\cite{commonvoice} version 16.1 and MUSAN~\cite{musan} noise sources, demonstrate that this framework significantly reduces the Word Error Rate compared to both the baseline system and text-only correction models. The experimental results confirm that explicit modeling of noise information as well as the combination of multiple hypotheses allow a generative LLM to achieve robust error correction even in severely degraded acoustic conditions. This work therefore provides a practical and scalable path toward noise-robust ASR systems in low-resource languages.

\section{Related Work}

Automatic Speech Recognition (ASR) systems have made significant advancements in recent years, yet challenges persist, particularly in low-resource languages like Persian. Traditional error correction methods often rely on rule-based approaches or statistical language models, which may not effectively address the complexities of ASR errors in such languages.

Robatian et al.~\cite{GEC-RAG2025} introduced GEC-RAG, a novel framework that enhances ASR accuracy by integrating retrieval-augmented generation. This approach treats the ASR system as a black-box and utilizes a knowledge base constructed from ASR predictions and their corresponding ground truths to retrieve lexically similar examples. By providing relevant error patterns to a Generative Large Language Model (LLM), GEC-RAG enables targeted corrections, significantly reducing word error rates in Persian ASR systems.

Dashti et al.~\cite{PERCORE2024} proposed PERCORE, a deep learning-based framework for Persian spelling correction that incorporates phonetic analysis. By integrating phonetic representations into the learning process, PERCORE effectively corrects both non-word and real-word spelling errors, achieving high F1-scores in evaluations.

Li et al.~\cite{LA-RAG2024} developed LA-RAG, a retrieval-augmented generation approach designed to enhance LLM-based ASR accuracy. LA-RAG leverages fine-grained token-level speech datastores and a speech-to-speech retrieval mechanism to improve ASR performance, particularly under varied acoustic conditions such as accents.

Sedghiyeh et al.~\cite{PSRB2025} introduced the Persian Speech Recognition Benchmark (PSRB), a dataset designed to assess Persian ASR systems under diverse linguistic and acoustic conditions. PSRB provides valuable insights into ASR performance and identifies key error types, facilitating the development of more robust ASR models.

Ghosh et al.~\cite{DARAG2025} proposed DARAG, a domain-agnostic approach that combines data augmentation and retrieval-augmented generation to improve generative error correction for ASR systems. DARAG enhances the generalization of error correction models, particularly in out-of-domain scenarios.

Recent research has also explored multilingual LLMs for direct 1-best ASR output correction. Studies have fine-tuned a single multilingual LLM covering over 100 languages to correct errors in ASR outputs without requiring N-best lists. Results show that even without N-best hypotheses, the approach significantly improves ASR performance, and transfer learning between languages with shared writing systems (e.g., Persian and Arabic) can enhance error correction in low-resource languages~\cite{li2024multilingual1best}.

GER~\cite{Ma2023AsrErrorCorrection} (Generative Error Correction) is a text-rewriting framework that formulates ASR post-processing as a generative sequence-to-sequence task. 
Instead of relying on handcrafted rules, edit-distance heuristics, or alignment-based correction, GER employs a generative model (e.g., a Transformer-based LLM) to rewrite the noisy 1-best ASR output into a clean and fluent transcription. 
By leveraging the model's ability to capture long-range contextual dependencies, GER can correct semantic, morphological, and phonetic errors that are difficult for traditional approaches to handle. 
GER has subsequently served as the foundation for extended variants such as Denoising GER~\cite{Liu2025DenoisingGER} and RobustGER~\cite{hu2024llmnoiseasr}.

Hu et al.~\cite{hu2024clozeGER} introduced ClozeGER, a framework using a multimodal LLM (SpeechGPT) that takes both the decoded text and the speech signal as input. By reformulating ASR correction as a cloze task and injecting speech logits, ClozeGER removes redundant inputs and achieves superior performance compared to standard ASR preprocessing with LLMs across nine benchmark ASR datasets.

Furthermore, Chen et al.~\cite{chen2023hyporadise} proposed HyPoradise, an open baseline for generative speech recognition using LLMs, highlighting the potential of large language models to generalize across diverse ASR tasks. Similarly, Hu et al.~\cite{hu2024llmnoiseasr} demonstrated that LLMs can efficiently learn noise-robust speech recognition by introducing RobustGER, emphasizing their effectiveness in low-resource and noisy scenarios.

Collectively, these studies highlight a shift towards integrating advanced machine learning techniques, including large language models, retrieval-augmented generation, multimodal inputs, and phonetic analysis, to address ASR error correction challenges, particularly in low-resource languages.

\section{Methodology}

This study investigates the application of large language models (LLMs) for improving Automatic Speech Recognition (ASR) accuracy, with a particular emphasis on the Persian language. We propose a novel framework, \textbf{GER + Text Denoising (Ours)}, which leverages Error Level Noise (ELN) vectors as auxiliary inputs to enhance transcription quality by explicitly modeling noise within the language space. The overall architecture is illustrated in Figure~\ref{fig:block_diagram} (Panel~d), shown alongside other baseline and previous methods. For experimentation, the ASR outputs include the top-5 hypotheses (\textit{5-best list}) generated by the underlying recognition system.

Our approach builds upon the ideas of RobustGER~\cite{hu2024llmnoiseasr} (Figure~\ref{fig:block_diagram}, Panel~c) but introduces key modifications to achieve computational efficiency. While RobustGER incorporates Knowledge Distillation (KD) to capture Language Noise and may also perform Audio Noise Distillation, our method simplifies this process by focusing solely on \textbf{Text Noise}. Specifically, we extract sentence- and token-level ELN vectors to quantify semantic and structural inconsistencies across ASR hypotheses. Unlike RobustGER, our approach omits the Audio Noise Distillation and Mutual Information Neural Estimation (MINE) stages. The resulting ELN vectors are provided as conditioning inputs to the LLM, serving as a noise-aware signal that guides the model toward improved transcription accuracy. In essence, this acts as a form of ``Text Denoising’’ directly within the language representation space.

\subsection{System Architectures Overview}

Figure~\ref{fig:block_diagram} illustrates the four system architectures used for ASR noise Robustness.
Panels (1, a) through (1, c) are adapted from Hu et al. ~\cite{hu2024llmnoiseasr}, showing: (a) GER ~\cite{chen2023hyporadise}, \cite{Yang2023GenerativeSR} — baseline
grammatical error revision model applied directly to noisy ASR outputs; (b) GER + Audio Denoising ~\cite{Liu2025DenoisingGER} — incorporation of audio-level noise reduction before
ASR transcription; (3) RobustGER \cite{hu2024llmnoiseasr} — enhancement of robustness through language-space noise modeling and knowledge distillation (KD); (4) GER + Text Denoising (Ours) — the proposed method in this paper, which introduces text-level noise embeddings (Text Noise) to denoise ASR hypotheses in the language space,
yielding improved correction accuracy across both English and Persian data.

\subsection{Hypothesis Generation}

\begin{figure*}[ht]
\centering
\includegraphics[width=\textwidth]{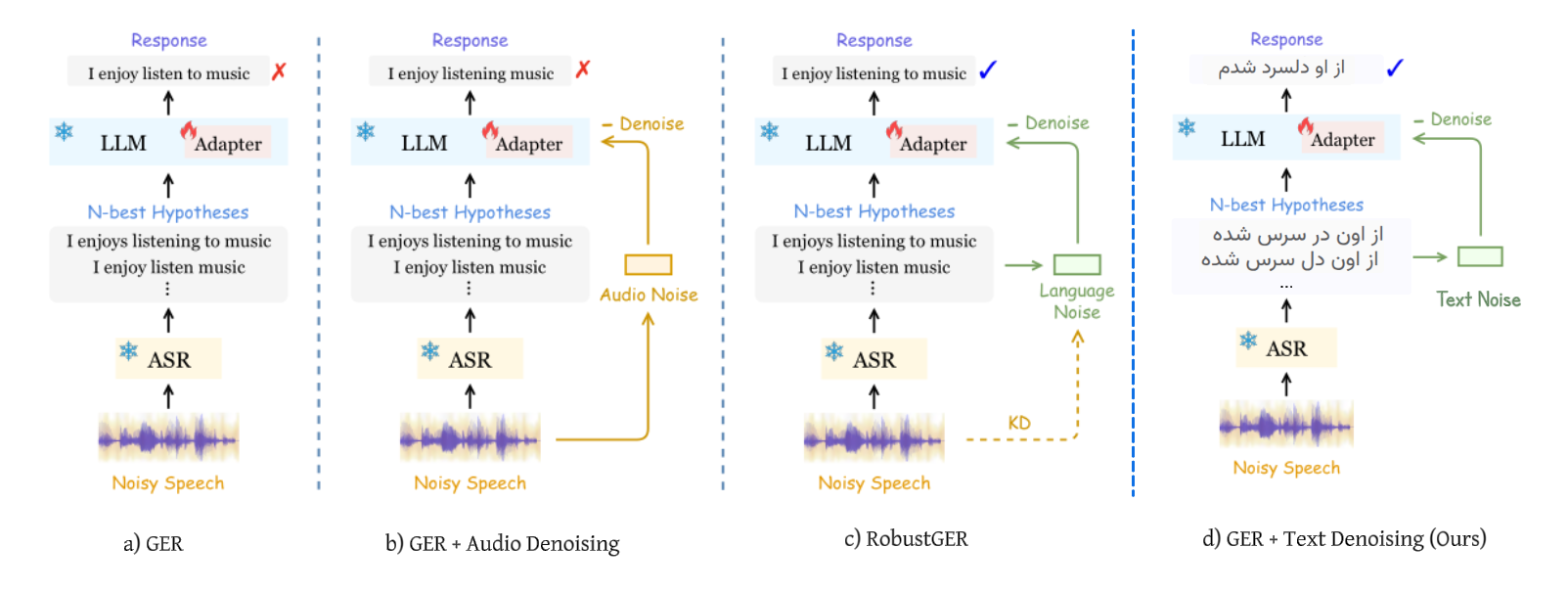}
\caption{Overview of a) GER~\cite{chen2023hyporadise} b) GER + Audio Denoising~\cite{Liu2025DenoisingGER} c) RobustGER~\cite{hu2024llmnoiseasr} and d) GER + Text Denoising(Ours)}
\label{fig:block_diagram}
\end{figure*}

As illustrated in Figure~\ref{fig:block_diagram}, the process begins with a Noisy Speech input fed into the ASR system, which produces an N-best list of transcription hypotheses. To build a high-quality dataset for ASR error correction, we utilized the Whisper~\cite{radford2023whisper} large model to generate multiple candidate transcriptions for each audio sample. To further reduce WER, a fine-tuned Persian variant of Whisper\footnote{https://huggingface.co/vhdm/whisper-large-fa-v1}
 was employed. For each input audio file, we applied beam search decoding to generate an n-best list, retaining up to five unique hypotheses per sample. In cases where fewer than five hypotheses were available, existing transcriptions were randomly duplicated to maintain a consistent list length ($n=5$).

All hypotheses and ground-truth transcriptions underwent a comprehensive Persian text normalization pipeline to ensure uniform formatting. The preprocessing steps included: (i) converting small numbers to Persian words while keeping larger numerical values intact, (ii) standardizing Unicode characters, (iii) correcting spacing and half-spacing errors, (iv) converting English digits to Persian digits, (v) removing punctuation marks, and (vi) eliminating extra whitespace. This normalization process guarantees consistency across samples and enables accurate evaluation of transcription performance using the Word Error Rate (WER) metric.

The resulting dataset provides a robust foundation for training and evaluating ASR error correction models. By incorporating diverse hypothesis variations, it effectively captures the range of possible transcription errors encountered in real-world scenarios—an essential component for reliable ELN vector computation in subsequent modeling stages.

\subsection{ELN Vector Extraction}

For a given audio sample, let $\mathcal{H} = \{H_1, H_2, \dots, H_n\}$ denote the set of $n$ ASR hypotheses generated by the system (here $n=5$). The ELN vector consists of two components: sentence-level ($\mathbf{v}_{\text{sent}}$) and token-level ($\mathbf{v}_{\text{tok}}$). This vector effectively models the Text Noise shown in Figure \ref{fig:block_diagram} (Panel d) and is fed back into the LLM-Adapter framework~\cite{zhang2023llmadapter}. The computation of two components, final ELN vector and also magnitude of the ELN vector are explained below:

\paragraph{Sentence-level ELN}

Each hypothesis $H_i$ is embedded into a fixed-dimensional semantic vector of dimension $d$, using a sentence embedding model (e.g., Sentence-BERT~\cite{reimers-2019-sentence-bert}):

\[
\mathbf{e}_i = \text{Embed}_{\text{sent}}(H_i) \in \mathbb{R}^d
\]

The sentence-level ELN vector is then computed as the mean pairwise difference of hypothesis embeddings:

\[
\mathbf{v}_{\text{sent}} = \frac{2}{n(n-1)} \sum_{i=1}^{n} \sum_{j=i+1}^{n} (\mathbf{e}_i - \mathbf{e}_j)^2
\]

where $(\cdot)^2$ denotes element-wise squaring. This captures semantic variance among all hypotheses.

\paragraph{Token-level ELN}

Each hypothesis is tokenized: $H_i = [t_{i,1}, t_{i,2}, \dots, t_{i,L_i}]$, and padded to the maximum length $L_{\max} = \max_i L_i$. Tokens are embedded into vectors of dimension $d'$:

\[
\mathbf{t}_{i,k} = \text{Embed}_{\text{tok}}(t_{i,k}) \in \mathbb{R}^{d'}\]

The token-level ELN vector is computed as the mean pairwise squared difference at each token position:

\[
\mathbf{v}_{\text{tok}} = \frac{1}{L_{\max}} \sum_{k=1}^{L_{\max}} \frac{2}{n(n-1)} \sum_{i=1}^{n} \sum_{j=i+1}^{n} (\mathbf{t}_{i,k} - \mathbf{t}_{j,k})^2\]

\paragraph{Final ELN Vector}

Finally, sentence- and token-level vectors are concatenated to form the final ELN vector:

\[
\mathbf{v}_{\text{ELN}} = \mathbf{v}_{\text{sent}} \, \Vert \, \mathbf{v}_{\text{tok}} \in \mathbb{R}^{d + d'}\]

This vector encodes multi-level linguistic noise information and is used as an additional input to the LLM for error correction, as indicated by the Text Noise input in the final panel of Figure \ref{fig:block_diagram}.

\paragraph{ELN Magnitude}
For analysis purposes, we define the \emph{magnitude} of an ELN vector as its $L_2$ norm:

\[
\|\mathbf{v}_{\text{ELN}}\|_2 = \sqrt{\sum_{i=1}^{d+d'} (\mathbf{v}_{\text{ELN},i})^2}.
\]

Here, $\mathbf{v}_{\text{ELN},i}$ denotes the $i$-th feature of the $\mathbf{v}_{\text{ELN}}$ vector. This scalar value provides a single measure of the overall noise-induced disagreement among hypotheses, with higher values indicating greater linguistic variability and potential transcription difficulty.

\section{Experimental Setup}

To evaluate the effectiveness of the proposed approach, we conducted three main experiments, each following the same prompt template designed to guide the large language model (LLM) in performing transcription error correction:

\textit{``You are a transcription error correction assistant and linguistics expert specializing in improving transcriptions produced by Automatic Speech Recognition (ASR) systems. Your task is to perform error correction based on the words in top 5 hypotheses generated by the ASR system. You can also correct the sentences yourself based on their meaning, ensuring spelling is correct. Do not use synonyms. Analyze linguistic context and provide corrected ASR hypothesis directly.''}

\subsection{Datasets}

Two publicly available corpora were used to construct the noisy Persian ASR dataset: one providing clean speech data and another supplying noise sources.

\paragraph{Common Voice 16.1 (Persian)}
This corpus served as the source of clean speech samples. Common Voice, developed by Mozilla, is a large-scale, multilingual, crowdsourced speech dataset. The Persian subset (version 16.1) includes over 40,000 validated audio recordings, corresponding to roughly 90 hours of transcribed speech. Each file is sampled at $16\ \text{kHz}$ and features diverse speakers varying in age, gender, and regional accent, contributing to the linguistic richness and representativeness of the dataset used in this study.

\paragraph{MUSAN}
To simulate realistic acoustic conditions, we employed the MUSAN corpus (Music, Speech, and Noise), a widely used resource for training and evaluating models in voice activity detection and noise-robust ASR. Specifically, the ambient noise subset was used, which comprises approximately 930 WAV files, totaling around 6 hours of audio.

The noise types in this subset are highly diverse, encompassing environmental sounds such as rain, thunder, wind, crowd chatter, footsteps, and paper rustling, as well as artificial noises like phone dial tones.
These diverse, non-speech noise samples were essential for constructing a realistic and challenging noisy evaluation environment that mirrors everyday acoustic conditions.

\paragraph{Data Augmentation and Noisy Speech Generation}
To enhance the robustness of the error correction model against various acoustic distortions, we synthetically generated noisy speech data by introducing controlled noise into clean audio samples from the Common Voice dataset.

Each noisy utterance, denoted as $\mathbf{y}$, was produced by combining a clean speech signal $\mathbf{x}_{\text{clean}}$ with a randomly selected noise segment $\mathbf{n}$ (sourced from the MUSAN corpus or generated Gaussian noise) at a specified signal-to-noise ratio (SNR) level, as expressed by:
$$\mathbf{y} = \mathbf{x}_{\text{clean}} + \mathbf{n}_{\text{scaled}}$$

For each clean sample, we applied a mixture of noise types to create realistic and variable acoustic conditions. Specifically, a randomly selected MUSAN ambient noise segment was added at an $\text{SNR}$ sampled between $0\ \text{dB}$ and $15\ \text{dB}$. In addition, additive white Gaussian noise was introduced, with its amplitude randomly drawn from the range $[0.001, 0.015]$.

This carefully designed augmentation process ensured the resulting dataset encompassed a broad range of noise intensities and characteristics. As a result, it provided a challenging and diverse evaluation environment, enabling a more comprehensive assessment of the proposed noise-aware ASR correction models.
\subsection{Evaluation Metrics}

Model performance was evaluated using Word Error Rate (WER), computed as:

\[
\text{WER} = \frac{S + D + I}{N} \times 100\%
\]

where $S$ is the number of substitutions, $D$ deletions, $I$ insertions, and $N$ total words in the reference transcription. 

\paragraph{Noise Types and SNR Conditions}
To ensure a controlled and interpretable evaluation of robustness, we report results under four noise conditions: Clean, Mixed, SNR = 5~dB, and SNR = 10~dB.

\textbf{Noise Types.} All noisy conditions are generated from the MUSAN ambient noise subset and additive white Gaussian noise. MUSAN includes a diverse set of environmental sounds such as rain, wind, footsteps, crowd chatter, cafeteria ambience, paper rustling, and mechanical background noise. Gaussian noise models sensor-level acoustic corruption.

\textbf{Signal-to-Noise Ratio (SNR).} SNR is defined as the ratio between the power of the clean speech ($P_{\text{speech}}$) and the power of the added noise ($P_{\text{noise}}$):
\[
\text{SNR (dB)} = 10 \log_{10} \left( \frac{P_{\text{speech}}}{P_{\text{noise}}} \right).
\]
Lower SNR values indicate more severe noise corruption.

\begin{itemize}
    \item \textbf{Clean}: No noise added. (5000 samples)
    \item \textbf{Mixed Noise}: Each utterance is corrupted with a randomly chosen MUSAN noise type at a randomly sampled SNR between 0–15~dB, plus low-amplitude Gaussian noise. (5000 samples)
    \item \textbf{SNR = 5 dB}: Each utterance is corrupted with a randomly chosen MUSAN noise type at a fixed SNR = 5~dB, plus low-amplitude Gaussian noise. (1000 samples)
    \item \textbf{SNR = 10 dB}: Each utterance is corrupted with a randomly chosen MUSAN noise type at a fixed SNR = 10~dB, plus low-amplitude Gaussian noise. (1000 samples)
\end{itemize}

These controlled SNR conditions allow us to compare performance across both (i) fixed, interpretable noise levels and (ii) a broad, real-world–like distribution (Mixed).

\subsection{Experiments}

We compare three settings:

\begin{enumerate}
    \item \textbf{Base Model (Zero-shot)}: LLaMA-2-7B~\cite{touvron2023llama2} without fine-tuning, directly correcting ASR outputs using the prompt.
    \item \textbf{Fine-tuned (Text-only)}: LLaMA-2-7B fine-tuned with LoRA~\cite{hu2021lora} on 5-ASR-hypothesis input without ELN vectors.
    \item \textbf{Fine-tuned + ELN (Ours)}: LLaMA-2-7B fine-tuned with LoRA incorporating ELN vectors. ELN vectors were mapped through a small MLP to match the LLM embedding dimension and prepended as prefix embeddings.
\end{enumerate}

\subsection{Setup}

All fine-tuning experiments used:
\begin{itemize}
    \item 4-bit quantization to reduce memory usage
    \item Custom DataCollator for batch padding
    \item Learning rate: $2\times10^{-4}$, 3 epochs
    \item Gradient accumulation and checkpointing
    \item Cosine scheduler with weight decay
\end{itemize}

This setup allows evaluation of how ELN vectors improve the model's ability to correct ASR errors in noisy and clean speech.

\section{Results and Discussion}

\subsection{Quantitative Results}

Table~\ref{table:quantitative-results} reports the Word Error Rate (WER, \%) of all evaluated models under four noise conditions: Clean, Mixed Noise (covering 0–15 dB SNR plus Gaussian noise), SNR = 5 dB, and SNR = 10 dB. The Persian noisy dataset was constructed from Common Voice 16.1 and augmented with MUSAN and Gaussian noise. For reference, the Raw Whisper baseline follows Radford et al.~\cite{radford2023whisper}, while the Base Model (Zero-shot) corresponds to LLaMA-2~\cite{touvron2023llama2}. All WER values are expressed in percentage.

\vspace{1em}

\begin{table*}[h]
\centering
\caption{WER (\%) comparison of Raw Whisper, Base LLaMA2 (zero-shot), LLaMA2 fine-tuned without ELN, and the ELN-conditioned model across four acoustic conditions.}
\label{table:quantitative-results}
\large % <-- COMMAND TO MAKE THE FONT LARGER
\begin{tabular}{@{} l S[table-format=2.2] S[table-format=2.2] S[table-format=2.2] S[table-format=2.2] @{}}
    \toprule
    \textbf{Method} & {\textbf{Clean}} & {\textbf{Mixed Noise}} & {\textbf{SNR = 5 dB}} & {\textbf{SNR = 10 dB}} \\
    \midrule
    Raw Whisper & 24.80 & 31.10 & 42.70 & 38.30 \\
    Base Model (Zero-shot) & 62.43 & 64.58 & 70.63 & 67.75 \\
    Fine-tuned (No ELN) & 24.06 & 30.79 & 39.76 & 31.59 \\
    \midrule
    \textbf{Fine-tuned + ELN (Ours)} & \textbf{24.39} & \textbf{24.84} & \textbf{32.34} & \textbf{28.02} \\
    \bottomrule
\end{tabular}
\end{table*}

\subsection{Comparison with RobustGER on VB-DEMAND}

To further assess the noise robustness and cross-lingual adaptability of our proposed approach, we performed a comparative evaluation against the state-of-the-art RobustGER model~\cite{hu2024llmnoiseasr} using a standard English benchmark dataset.

For this comparison, we employed the VoiceBank-DEMAND (VB-DEMAND) corpus, a well-established dataset for supervised speech enhancement and noise-robust ASR evaluation. This dataset is created by combining clean English speech recordings from the VoiceBank (VCTK) corpus~\cite{veaux2017vctk} with diverse real-world and synthetic noise samples sourced from the DEMAND database~\cite{thiemann2013demand}. As a result, it provides paired clean and noisy utterances, making it an ideal benchmark for assessing systems designed to handle acoustic degradation.

In particular, we used the $\texttt{VoiceBank-DEMAND-16k}$ variant hosted on Hugging Face\footnote{\url{https://huggingface.co/datasets/JacobLinCool/VoiceBank-DEMAND-16k}}
. This version includes 11,572 examples of paired clean and noisy speech in the training set and 824 samples in the test set, featuring noise conditions such as bus, cafe, and office environments across varying SNR levels.

For the English ASR baseline, we utilized the $\texttt{small.en}$ variant of the Whisper model to generate N-best hypotheses. Our fine-tuned LLaMA2-7B model, equipped with ELN conditioning was then trained on the VB-DEMAND training split and evaluated on the official test set. The results of this comparison are presented in Table~\ref{table:vb-demand-comparison}.

\begin{table}[h]
\centering
\caption{Comparison of WER (\%) between the proposed ELN-conditioned model and RobustGER~\cite{hu2024llmnoiseasr} on the \textbf{VB-DEMAND dataset}.}
\label{table:vb-demand-comparison}
\begin{tabular}{@{} l S[table-format=2.2] S[table-format=1.2] @{}}
    \toprule
    \textbf{Method} & {\textbf{Baseline WER (\%)}} & {\textbf{Improved WER (\%)}} \\
    \midrule
    RobustGER  & 13.00 & 10.70 \\
    \textbf{Ours (Fine-tuned + ELN)} & 7.93 & \textbf{3.96} \\
    \bottomrule
\end{tabular}
\end{table}
Our model achieved a significantly lower final WER of $3.96\%$ , outperforming the $10.70\%$ reported for RobustGER. This substantial improvement highlights the effectiveness of incorporating ELN-based noise representations for resilient ASR error correction under challenging acoustic conditions. Furthermore, the strong performance on an English benchmark demonstrates that our proposed Text Denoising approach—originally developed for Persian—generalizes effectively across languages, confirming its cross-lingual robustness.

For clarity, in Table~\ref{table:vb-demand-comparison}, Baseline WER refers to the Word Error Rate of the Whisper ASR system prior to any correction, while Improved WER indicates the WER after applying the respective error correction model.

\subsection{Error Analysis}

To better understand the behavior of our model, we analyzed the relationship between the magnitude of ELN vectors (measured by the \(L_2\) norm) and the corresponding Word Error Rate (WER). Our findings reveal that most samples exhibit relatively small ELN magnitudes (below 20), which correlate with low and stable WER values. However, as ELN magnitude increases, WER becomes more variable, and extreme transcription errors occur more frequently.

By categorizing samples according to ELN magnitude ranges, we observed that lower magnitudes correspond to smaller median WER and tighter distributions, indicating consistent performance. In contrast, higher magnitudes are associated with increased median WER, greater variance, and a higher frequency of outlier cases. This trend suggests that stronger linguistic noise leads to more unstable transcriptions and an elevated likelihood of significant recognition errors.

Overall, these observations underscore the strong correlation between ELN vector magnitude and ASR reliability. Low ELN values are indicative of stable and accurate recognition, whereas higher magnitudes signal increased uncertainty and degradation in performance. These insights reinforce the value of ELN-based conditioning as a mechanism for mitigating noise-induced variability and improving the resilience of ASR correction models under challenging acoustic environments.

\subsection{Key Findings}

Our experiments revealed several noteworthy insights. First, task-specific fine-tuning proved essential for achieving accurate Persian ASR correction. Beyond its effectiveness in the target language, the ELN-conditioned model also exhibited strong cross-lingual generalization, attaining competitive results on the English VB-DEMAND benchmark.

Second, the integration of ELN vectors with textual inputs significantly reduced the Word Error Rate (WER), especially under moderate noise levels such as $5\ \text{dB}$ SNR. Under these conditions, greater disagreement among ASR hypotheses provides richer information for the ELN representation, allowing the model to perform more noise-aware correction.

Finally, analysis of the ELN magnitude—quantified by its $L_{2}$ norm—showed a clear positive correlation with WER and performance variability across samples. This relationship indicates that ELN magnitude effectively reflects the difficulty or noise sensitivity of each utterance, serving as a reliable, model-agnostic indicator of recognition uncertainty.

\subsection{Limitations}

Despite the encouraging results, several limitations of this study should be acknowledged. First, the experiments were conducted using the Common Voice (Persian) dataset augmented with synthetic MUSAN noise, which may not fully represent the diversity and complexity of real-world Persian speech, particularly in conversational, dialectal, or domain-specific contexts. Second, the evaluation primarily relied on the Word Error Rate (WER) metric. While WER is a useful quantitative measure, future work should incorporate human-based evaluations or semantic similarity metrics to more comprehensively assess improvements in fluency, coherence, and meaning preservation. Finally, the current approach models noise solely through textual proxies using Error Level Noise (ELN) vectors, without leveraging direct acoustic evidence. Integrating explicit acoustic features in future research could provide a more complete understanding of noise influence and further enhance the robustness of language-space noise modeling.

% \subsection{Future Work}

% Building on the success of the ELN-conditioned framework, we outline three promising directions for future research. 

% First, we plan to develop a bimodal LLM adapter that integrates both the language-space ELN vectors and direct acoustic representations (e.g., from a frozen Whisper encoder or a speech enhancement model). This extension would enable a genuinely multimodal understanding of noise, addressing the current limitation of excluding explicit audio information during the correction phase. 

% Second, we aim to extend the model to condition its generation on preceding corrected utterances and dialogue context. This is particularly important for conversational Persian—a highly contextual language—as it ensures semantic coherence and handles long-range dependencies beyond single-utterance correction. 

% Finally, to translate these academic findings into deployable applications, future work must investigate real-time ASR correction. Since current LLM-based post-processing, relying on N-best hypotheses and complex ELN computations, introduces latency, potential solutions include model quantization, knowledge distillation from the LLaMA-2 model, or implementing streaming inference techniques that process partial hypotheses incrementally for near-instantaneous correction in live speech systems.

\section{Conclusion}

This study explored the use of large language models (LLMs) for post-hoc correction of Automatic Speech Recognition (ASR) outputs, focusing primarily on the low-resource Persian language. We proposed a novel and computationally efficient framework which conditions a fine-tuned LLaMA-2-7B model on an Error Level Noise (ELN) vector derived from the n-best ASR hypotheses.

By explicitly modeling linguistic noise within the language space using ELN, our approach substantially enhances transcription robustness without requiring modifications to the original ASR architecture. In summary, conditioning LLMs with structured noise representations provides a practical and scalable approach to improving ASR reliability, especially for low-resource languages operating under noisy conditions.

% References to key methods and models referenced in the conclusion:
% \cite{touvron2023llama2, radford2023whisper, hu2024llmnoiseasr}

% \section*{Acknowledgment}
% Optional acknowledgment section.

\bibliographystyle{IEEEtran}
\bibliography{references}  % references.bib file

\end{document}